%% file: ms.tex
\documentclass{article}
\usepackage[preprint]{spconf}
\usepackage{times}
\usepackage{epsfig}
\usepackage{graphicx}
\usepackage{graphics}
\usepackage{wrapfig}
\usepackage{amsmath}
\usepackage{amssymb}
\usepackage{soul, color}
\usepackage{url}            % simple URL typesetting
\usepackage{booktabs}       % professional-quality tables
\usepackage{physics}        % tons of useful math typesetting commands
\usepackage{amsfonts}       % blackboard math symbols
\usepackage{nicefrac}       % compact symbols for 1/2, etc.
\usepackage{microtype}      % micro-typography
\usepackage{enumerate}      % better enumerations
\usepackage{graphicx}
\usepackage{hyperref}

\usepackage{fancyhdr} %To use headertext.
\usepackage{caption}
\usepackage{subcaption}

\newcommand{\figr}{Fig.~\ref}				% figure ref
\newcommand{\tabref}{Tab.~\ref}				% table ref
\newcommand{\Sec}{Sec.~\ref}					% section ref
\begin{document}
% To add the credit to IEEE 
\fancypagestyle{firstpage}
{
    \fancyhead[L]{\tiny{© 20222 IEEE. Personal use of this material is permitted. Permission from IEEE must be obtained for all other uses, in any current or future media, including reprinting/republishing this material for advertising or promotional purposes, creating new collective works, for resale or redistribution to servers or lists, or reuse of any copyrighted component of this work in other works.}}    
}

\thispagestyle{firstpage}
%%%%%%%%% TITLE
\title{Simpler is better: spectral regularization and up-sampling techniques for variational autoencoders}
% \footnotesize{
\name{Sara Björk$^1$, Jonas Nordhaug Myhre$^{2,1}$ and Thomas Haugland Johansen$^{2,1}$}
\address{
\small{
$^1$UiT The Arctic University of Norway,
$^2$NORCE Norwegian Research Centre}
}
% }

\maketitle
%%%%%%%%% ABSTRACT
\begin{abstract}
Full characterization of the spectral behavior of generative models based on neural networks remains an open issue.
Recent research has focused heavily on generative adversarial networks and the high-frequency discrepancies between real and generated images. 
The current solution to avoid this is to either replace transposed convolutions with bilinear up-sampling or add a spectral regularization term in the generator.
It is well known that Variational Autoencoders (VAEs) also suffer from these issues.

In this work, we propose a simple 2D Fourier transform-based spectral regularization loss for the VAE and show that it can achieve results equal to, or better than, the current state-of-the-art in frequency-aware losses for generative models.
In addition, we experiment with altering the up-sampling procedure in the generator network and investigate how it influences the spectral performance of the model.
We include experiments on synthetic and real data sets to demonstrate our results.

\end{abstract}

\section{Introduction}
\label{sec:Introduction}
\input{body/introduction.tex}

%%%%%%%%%%%%%%%%%%%%%%%%%%%%%%%%%%%%%%%%%%%%%%%%%%%
\subsection{Related Work}
\label{sec:background}
\input{body/background.tex}
%%%%%%%%%%%%%%%%%%%%%%%%%%%%%%%%%%%%%%%%%%%%%%%%%%%
%%%%%%%%%%%%%%%%%%%%%%%%%%%%%%%%%%%%%%%%%%%%%%%%%%%
\section{Methodology}
\label{sec:method}
\input{body/method.tex}
%%%%%%%%%%%%%%%%%%%%%%%%%%%%%%%%%%%%%%%%%%%%%%%%%%%
%%%%%%%%%%%%%%%%%%%%%%%%%%%%%%%%%%%%%%%%%%%%%%%%%%%
\section{Experiments}
\label{sec:exp}
\input{body/experiments}
%%%%%%%%%%%%%%%%%%%%%%%%%%%%%%%%%%%%%%%%%%%%%%%%%%%
%%%%%%%%%%%%%%%%%%%%%%%%%%%%%%%%%%%%%%%%%%%%%%%%%%%
\section{Conclusion and future work}
\label{sec:conclusion}
\input{body/conclusion.tex}
%%%%%%%%%%%%%%%%%%%%%%%%%%%%%%%%%%%%%%%%%%%%%%%%%%%
\section{Acknowledgements}
\label{sec:ack}
We thank Stian Normann Anfinsen at NORCE and Robert Jenssen at UiT for their valuable feedback. This work was financially supported by the Research Council of Norway (RCN), through its Centre for Research-based Innovation funding scheme (Visual Intelligence, grant no. 309439), and  Consortium  Partners.
%%%%%%%%%% References %%%%%%%%%%%%%%%%%%%%%%%%%%%%%

{\footnotesize
\bibliographystyle{IEEEbib}
\bibliography{refMend2.bib}
}

\end{document}

%% file: body/introduction.tex
\begin{figure}[tp]
\centering
 \resizebox{.8\linewidth}{!}
 {
\begin{minipage}{\columnwidth}
\includegraphics[width=0.30\columnwidth]{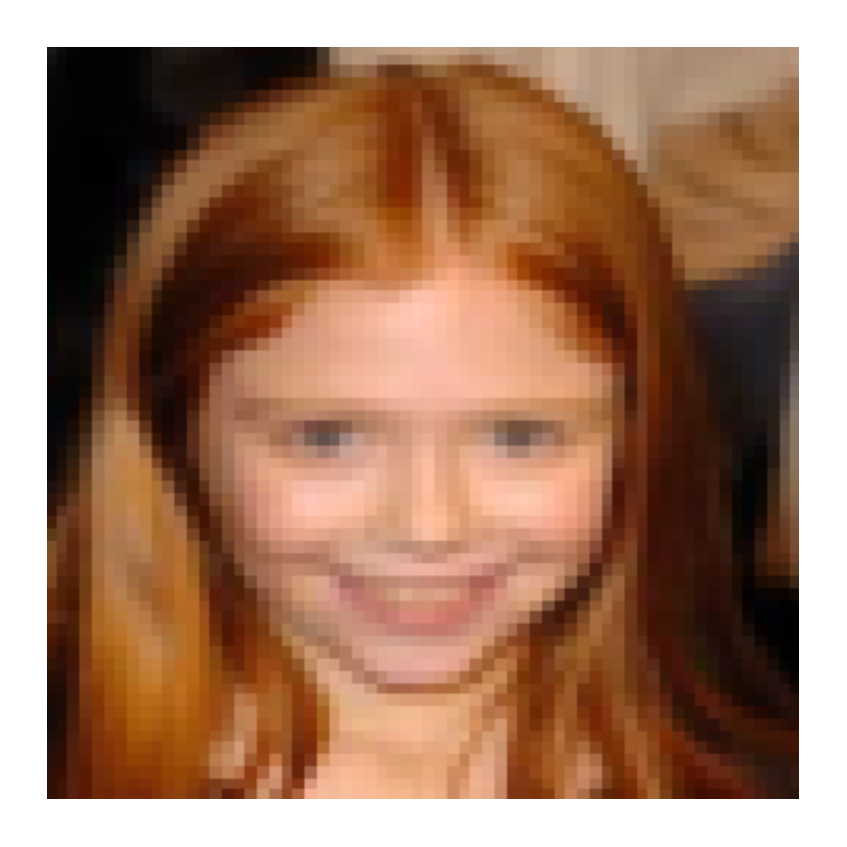}
\hspace*{-0.3cm}
\includegraphics[width=0.30\columnwidth]{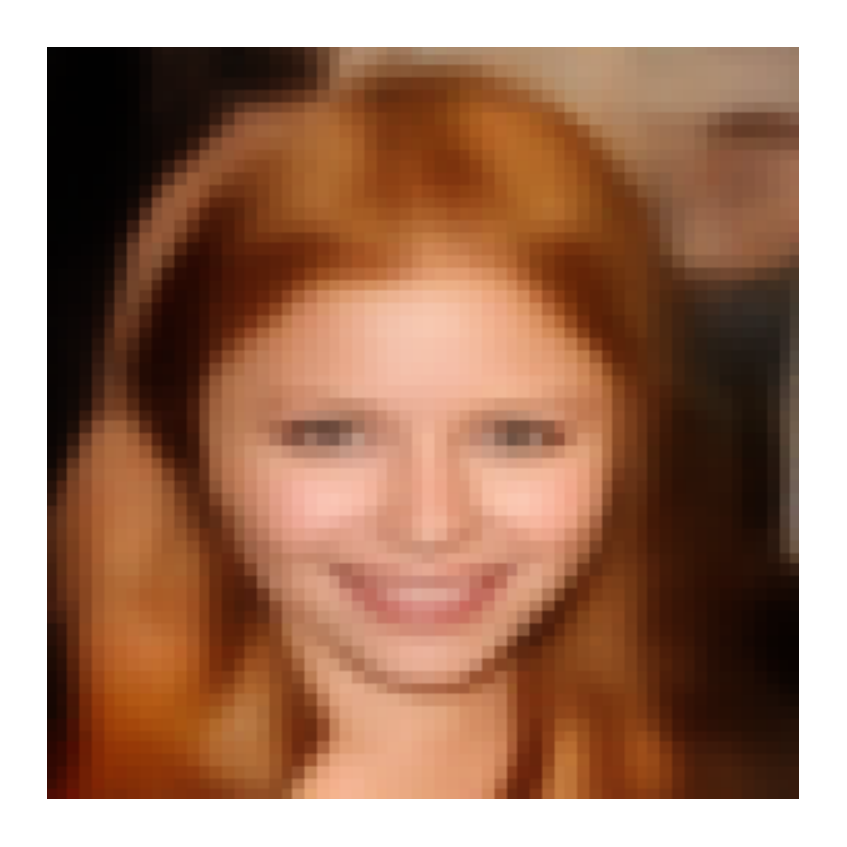}
\hspace*{-0.3cm}
 \includegraphics[width=0.30\columnwidth]{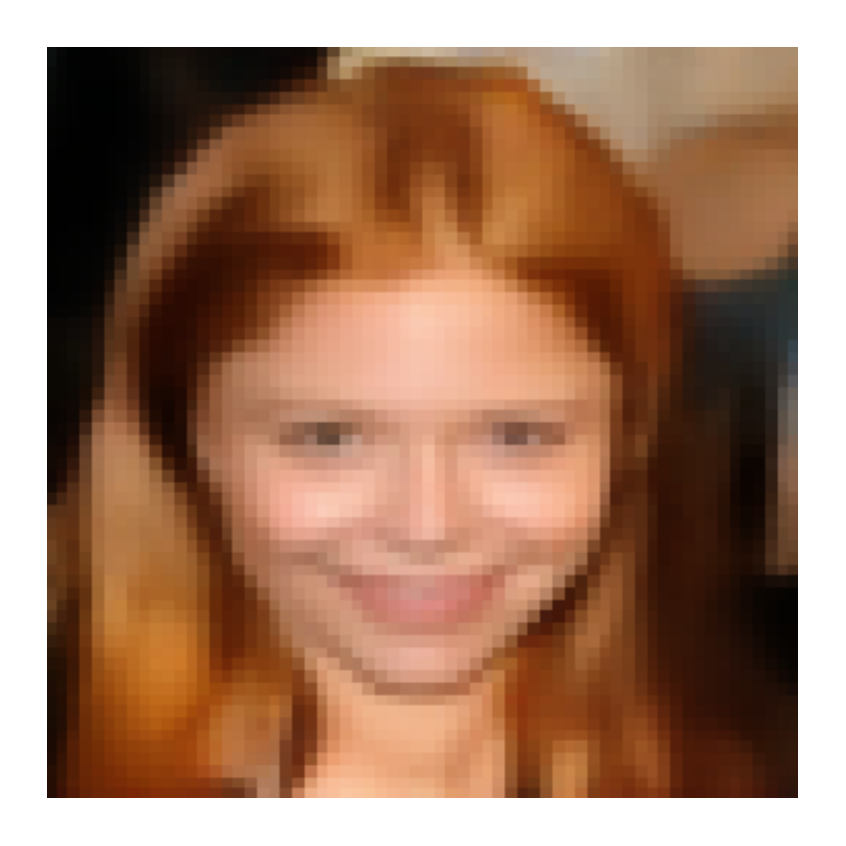}
\vspace*{-.2cm}
\\
%%%%%%%%%%%%%%%%%%%%%%%%%%%%%%%%%%%%%%%%%%%%%%%%%%%%%%%%%%%%%%%%%%%%%%%%%%%%
\subfloat[\label{fig:real}]{%
\includegraphics[width=0.30\columnwidth]{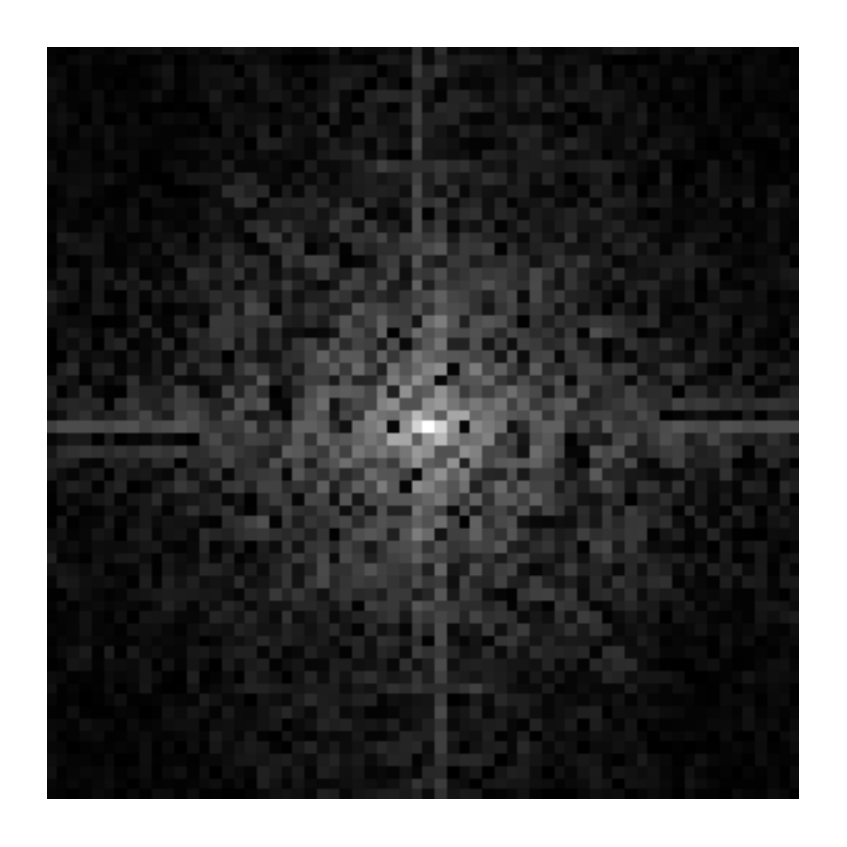}}
\hspace*{-0.2cm}
\subfloat[\label{fig:bce}]{%
\includegraphics[width=0.30\columnwidth]{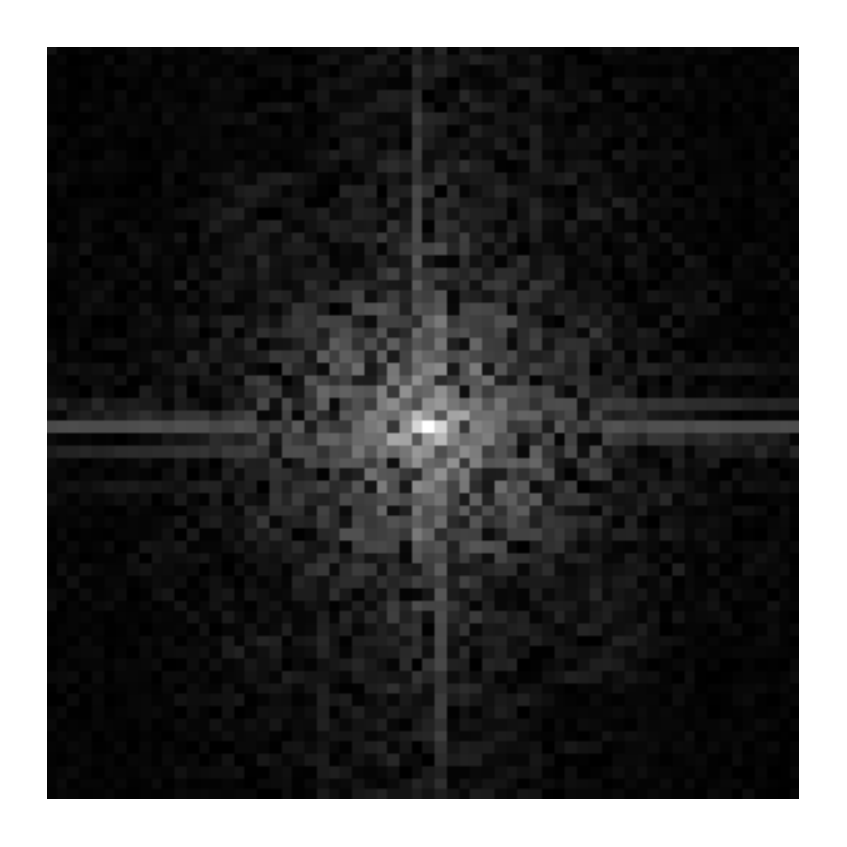}}
\hspace*{-0.2cm}
 \subfloat[\label{fig:fft}]{%
\includegraphics[width=0.30\columnwidth]{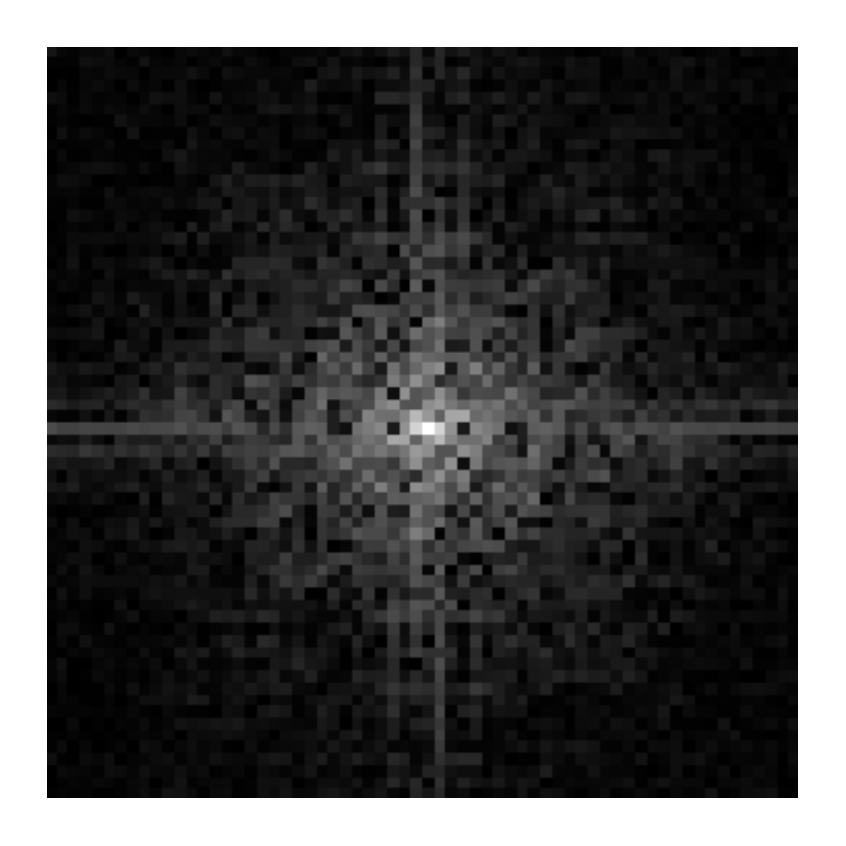}}
\\
\end{minipage}
}
\caption{
A sample from the CelebA dataset~\cite{Liu2015celeba}. Top row: \textbf{(a)}: real image, \textbf{(b)}: Vanilla VAE reconstruction and \textbf{(c)}: reconstruction from VAE with spectral regularization. Bottom row: FFT spectrum of the corresponding images. We note the discrepancies at the highest frequencies of the 2D Fourier spectrum in (b), compared to (a), and the lack of details in the spatial representation of the image. A simple 2D FFT regularization (c) achieves less blurriness in the spatial domain and less discrepancies in the Fourier spectrum. Figure is best viewed online.}
 \label{fig:samples}
\end{figure}

Generative deep neural network models such as the Generative Adversarial Network (GAN) \cite{Goodfellow2014GenerativeAdversarialNets} and the Variational Autoencoder (VAE) \cite{Kingma2014AutoEncodingVariationalBayes} have in recent years gained a lot of attention in e.g. face generation \cite{Karras2019a, Karras2020a, Vahdat2020a}, image-to-image translation or style-transfer \cite{Isola2017Pix2pix, Choi2020StarGANv2, Zhu2017} tasks.
The wide applicability of generative models has fostered a large body of research that aims to improve generative network architectures to enhance the quality of the generated images.
Most of this work has focused on proposed variations of spatial loss terms in the objective functions, which has led to a multitude of different GAN and VAE architectures, see e.g. \cite{Radford2015DCGAN, Gulrajani2017WGAN-GP, Mao2017LSGAN, Zhu2017,Child2021, Larsen2016}.
Although current methods generate very realistic-looking natural images, see e.g. \cite{Karras2019a, Karras2020a, Vahdat2020a}, generative neural network models are in general not able to reproduce the spectral distribution of natural images adequately.
Generated images still suffer from blurriness and lack of sharp details.
This issue is illustrated in columns (a) and (b) of \figr{fig:samples}, where the first column, (a), shows an original sample from the CelebA dataset \cite{Liu2015celeba}, and (b) is a blurry reconstruction of the same sample image from a VAE trained with a traditional spatial objective function.

The lack of high-frequency content can be partially explained by the \emph{spectral bias} of neural networks~\cite{Rahman2019OntheSpectralBias}; neural networks prioritize low-frequency components of the data in the early stages of learning. 
A growing body of research has investigated these findings, see e.g. \cite{Khayatkhoei2020SpatialFrequencyBias, Chandrasegaran2021, Wang2020HighfrequencyComponent, Sitzmann2020ImplicitNeuralRepresentations}, and ways to utilize this in e.g.~deep-fake detection \cite{Frank2020a, Durall2020WatchYourUpconvCNN, Dzanic2020, Jung2021}.
Others propose different ways to resolve, work around, or reduce the effects of a bias towards the low-frequency components \cite{Durall2020WatchYourUpconvCNN, Czolbe2020WatsonsPercepturalModel, Dzanic2020, Chen2020a, Czolbe2020WatsonsPercepturalModel, Chandrasegaran2021, Wang2021TheFrequencyDiscrepancy}. 
Another partial explanation for the discrepancy in the frequency content of generated images is the transposed convolution operation used in the up-sampling components of generative models.
Durall et al.~\cite{Durall2020WatchYourUpconvCNN} argue that the transposed convolution operation is causing the models' inability to learn the high-frequency content of the data and propose to append \emph{spectral regularization} (SR) to the spatial objective function to mitigate the effects caused by the up-sampling strategy. Others, see e.g. \cite{Chandrasegaran2021, Wang2021TheFrequencyDiscrepancy}, suggest to replace the last up-sampling operation in the architecture.

In this work we show that a simple frequency-aware loss that forces the generative model to focus on agreement of the overall spectral content of the data is equally effective, and sometimes better than the current state-of-the-art-in SR~\cite{Durall2020WatchYourUpconvCNN, Czolbe2020WatsonsPercepturalModel}.
Furthermore, we consider the effects of replacing the up-sampling operation in the last layer, similar to \cite{Chandrasegaran2021, Wang2021TheFrequencyDiscrepancy}. This allows us to evaluate the impact of the up-sampling operation both with and without spectral regularization. 
Concretely, we propose to incorporate a simple 2D Fourier transform agreement loss term to the overall objective function.
With this additional spectral agreement term, we wish to align the high-frequency components of the Fourier spectrum while penalizing unilateral learning of low-frequency components of the data. 
A comparison of the vanilla VAE in column (b) with the proposed loss in (c) of \figr{fig:samples} shows that the added SR term results in better spatial and spectral agreement with the original data. 
We empirically evaluate the SR loss term on synthetic and real datasets, and compare with two more complex SR methods; the azimuthal integration loss by Durall et al.~\cite{Durall2020WatchYourUpconvCNN} and the Watson perceptual loss from Czolbe et al.~\cite{Czolbe2020WatsonsPercepturalModel}. 
All three SR methods are compared against the baseline VAE objective function with its binary cross-entropy (BCE) loss.

The rest of the paper is organized as follows: we review some related work that focuses on spectral reconstruction with generative models in \Sec{sec:background}.
In \Sec{sec:method} we briefly introduce the VAE, SR with our 2D Fourier transform agreement loss, and introduce an alternative approach to the transposed convolution up-sampling operation.
Results from our experiments are presented and discussed in \Sec{sec:exp}.
Finally, \Sec{sec:conclusion} concludes this work with a summary of our most important findings.

%% file: body/background.tex
Many works have illustrated the problems of generative models and spectral reconstruction.
Several theories exist, but the most notable are \emph{spectral bias}~\cite{Rahman2019OntheSpectralBias,Khayatkhoei2020SpatialFrequencyBias} and issues related to the up-sampling operations in the final layers of the generator network~\cite{Durall2020WatchYourUpconvCNN,Wang2021TheFrequencyDiscrepancy}.
Karras et al.~\cite{Karras2020a} generate high-resolution images by first letting their network focus on low-resolution images and then progressively shift the training to consider higher-resolution images. However, as pointed out by Khayatkhoei and Elgammal~\cite{Khayatkhoei2020SpatialFrequencyBias}, application of the StyleGAN2 \cite{Karras2020a}, which samples at high frequencies, might avoid the spatial frequency bias without actually solving the issue: high-frequency components, such as sharp details, are not preserved to the same extent in data that has been sampled at very high resolution \cite{Khayatkhoei2020SpatialFrequencyBias}. Moreover, access to high-definition or high-resolution images is not always possible, especially not when working with e.g.~remote sensing data or medical data. Very deep architectures might also be unsuitable when considering available computational power or computation time in specific applications and projects.

There are numerous works in the last two years that either try to explain the frequency discrepancy from a theoretical perspective, such as \cite{Tabcik2020FourierFeaturesLetNetworks, Wang2020HighfrequencyComponent}, or acknowledge this drawback by proposing ways to resolve or reduce the effects of the spectral bias.
Particularly important is the work by Durall et al.~\cite{Durall2020WatchYourUpconvCNN}, which illustrates how standard up-sampling methods such as up-convolution or transposed convolution in the generator network result in generative models that are incapable of reproducing the spectral distribution of the data. 
While up-sampling methods lack high frequencies, transposed convolution, on the other hand, adds a large amount of high-frequency noise in the up-sampling process.
To overcome the issues the up-sampling process causes, they propose to include SR in the generator image-space-based loss.
The spectral part of the loss is a 1D representation of the Fourier power spectrum given by azimuthal integration over the radial frequencies.
In the same year, Czolbe et al.~\cite{Czolbe2020WatsonsPercepturalModel} also adopted the idea of SR by proposing a loss function based on Watson's visual perception model \cite{Watson1993DCTQuantizationMatrices}.
It mimics the human perception of image data using a weighted distance in frequency space and adjusting for contrast and luminance.
From now on, we refer to SR by azimuthal integration \cite{Durall2020WatchYourUpconvCNN} as the AI loss, and to \cite{Czolbe2020WatsonsPercepturalModel} as the Watson-DFT loss.
In \cite{Chandrasegaran2021}, Chandrasegaran et al.~argue that the spectral discrepancies are not inherent to the neural network, but an artifact from the up-sampling procedure.
They show promising results by replacing the last transposed convolution layer with either zero-insert scaling, nearest interpolation, or bilinear interpolation followed by traditional convolution.

%% file: body/method.tex
This section briefly introduces variational autoencoders and the proposed frequency-aware loss function used in this work.
A section describing the commonly used up-sampling procedures in convolutional neural networks is also included.

\subsection{Variational Autoencoders}
A variational autoencoder is a Bayesian generative model configured in an autoencoding fashion, with an encoder mapping the data, $x$, into a latent variable, $z$, and a decoder that maps the latent variable back to the original data space.
As usual in a Bayesian setup, the problem of inference is to find the posterior distribution $p(z | x)$.
Since the evidence $p(x)$ is typically intractable, a lower bound is optimized using variational inference~\cite{Kingma2014AutoEncodingVariationalBayes}: 
\begin{equation}
        \underset{\phi, \theta}{\mathrm{arg\,min}} \;\mathbb{E}_{q_{\phi}(z|x)}\left\{\log p_{\theta}(x|z)  \right\} - \beta \text{KL}\left(q_{\phi}(z|x)||p(z)\right).
\end{equation}
For full derivations, see \cite{Blei2017} or \cite{Kingma2019}.
Both $q_{\phi}(z|x)$ (encoder) and $p_{\theta}(x|z)$ (decoder) are modeled via neural networks. $p(z)$ is the prior over the latent variable $z$, which is commonly assumed to be multivariate Gaussian distributed.
Furthermore, identifying $\log p_{\theta}(x | z)$ as the negative BCE loss, we can replace this with an energy-based model, $p(x | z) \propto \exp{- L\qty(x, \mu_x\qty(z))}$ where $L$ is any function that leads to a proper probability density function~\cite{Czolbe2020WatsonsPercepturalModel}.
This formulation allows alternative reconstruction losses, such as the Watson perceptual loss used in~\cite{Czolbe2020WatsonsPercepturalModel}.

\subsection{A 2D frequency spectral regularization loss}
Czolbe et al. \cite{Czolbe2020WatsonsPercepturalModel} suggest that the combination of spectral and spatial components in the reconstruction loss helps improve the image quality of generated samples. 
Motivated by this, we propose a simple SR, the FFT loss, that combines a general spatial VAE loss with deviation measures of the real and imaginary components of the 2D Fourier transform.
In our frequency-aware loss function for the VAE, we replace $L$ with:
\begin{equation}
    \mathcal{L}\qty(x, \hat{x}) = \alpha\mathcal{L}_s\qty(x, \hat{x}) + (1-\alpha) \mathcal{L}_f\qty(x, \hat{x})
        \qcomma \alpha \in [0, 1].
        \label{eq:loss-full} \\
\end{equation}
\sloppy where $\mathcal{L}_s\qty(x, \hat{x})$ is the BCE loss computed in the spatial domain between the actual image, \textit{x}, and its reconstruction, $\hat{x}$. $\mathcal{L}_f\qty(x, \hat{x}) = \frac{1}{n}\sum \left(imag[\mathcal F \qty{x}]- imag[\mathcal F\qty{\hat x}]\right)^2 + $ $ \frac{1}{n}\sum \left(real[\mathcal F \qty{x}]- real[\mathcal F\qty{\hat x}]\right)^2$ is the MSE in the frequency domain, where $\mathcal F$ denotes the fast Fourier Transform.

\subsection{Up-sampling and transposed convolution}\label{sec:up}
To convert the latent vector into a higher-dimensional output space, e.g., transforming a low-dimensional Gaussian sample to an image, the generator needs to increase the resolution in each layer.
The most common strategy is to use the \emph{transposed convolution} operation, where the input is zero-padded and convolved with the appropriate filter.
See~\cite{Dumoulin2016} for a complete description.
The alternative approach is to split the transformation in two: \emph{up-sampling by interpolation} and \emph{convolution}.
The transposed convolution operation is known to have several shortcomings, such as high-frequency discrepancies and checkerboard artifacts~\cite{Chandrasegaran2021,Odena2016}. 
Chandrasegaran et al.~\cite{Chandrasegaran2021} propose multiple ways to perform the up-sampling in the last layer of a generator network.
Their results advise to use up-sampling with nearest-neighbor interpolation and a single convolutional block of kernel size $5 \times 5$.
We adopt this setup in this work, hereafter denoted 'N.1.5', and refer to \cite{Chandrasegaran2021} for a comprehensive evaluation of additional versions of the up-sampling procedure.

%% file: body/experiments.tex
We evaluate our proposed 2D spectral loss, i.e. the FFT loss, and compare its performance to the AI loss \cite{Durall2020WatchYourUpconvCNN}, the Watson-DFT loss \cite{Czolbe2020WatsonsPercepturalModel}, and the baseline spatial BCE loss (Vanilla-VAE).
In the case of RGB images, these are in \cite{Durall2020WatchYourUpconvCNN} first transformed to gray-scale before the AI loss is computed.
We performed additional experiments with the AI loss by separately computing it channel-wise on RGB images to evaluate whether the gray-scale transformation impacts its performance.
Our experiments show that the channel-wise AI loss performs similarly to the original AI loss, and it has been omitted in the reported results.
As both our proposed FFT loss and the Watson-DFT loss can be applied to both gray-scale and RGB images, there was no need to modify them for RGB images.

We employ three different datasets with increasing complexity for the evaluation: a simple gray-scale version of the Shape dataset by Jing et al.~\cite{jing2020implicitIRMAE}, the grey-scale MNIST dataset \cite{LeCun1998mnist}, and the RGB CelebA dataset \cite{Liu2015celeba} of celebrity faces.
The Shape and MNIST datasets were analyzed at 32x32 resolution, while the CelebA dataset was analyzed at 64x64 resolution. 
We choose to employ the simple VAE networks from \cite{jing2020implicitIRMAE}.
The focus of this work is to evaluate how SR, either alone or combined with last layer up-sampling \cite{Chandrasegaran2021}, can enhance image quality.
The influence of different network architectures on the models' ability to reproduce the high-frequency content of the data is beyond the scope of this work and has therefore been omitted.
The interested reader could consult \cite{Karras2019a, Karras2020a, Vahdat2020a} for examples of generative model architectures focused on generating high-resolution images from low-resolution images. 
To evaluate the performance of the models, we use the root mean squared error (RMSE) and azimuthal (polar coordinate) integration of the Fourier spectrum.
Since the AI loss focuses on alignment in the 1D representation of the Fourier power spectrum, the Vanilla-VAE on the alignment in the spatial domain, and our FFT loss on the alignment in the  2D representation of the Fourier spectrum, we choose to compute the RMSE in all these three domains. 
RMSE metrics for the Watson-DFT loss on the Shape dataset have been omitted from the reported results of \Sec{sec:exp1} and \Sec{sec:exp2}, since these models did not work correctly.
We argue that this could be an effect of the Shape dataset being too simple for a more complex loss, but did not investigate this further since the Shape dataset was included only to compare how different losses generalize from simple to more complex datasets.

\begin{figure}[t]
    \centering
    \vspace*{-0.15cm}
\subfloat[MNIST \label{fig:mnist}]{%
       \includegraphics[width=0.5\columnwidth]{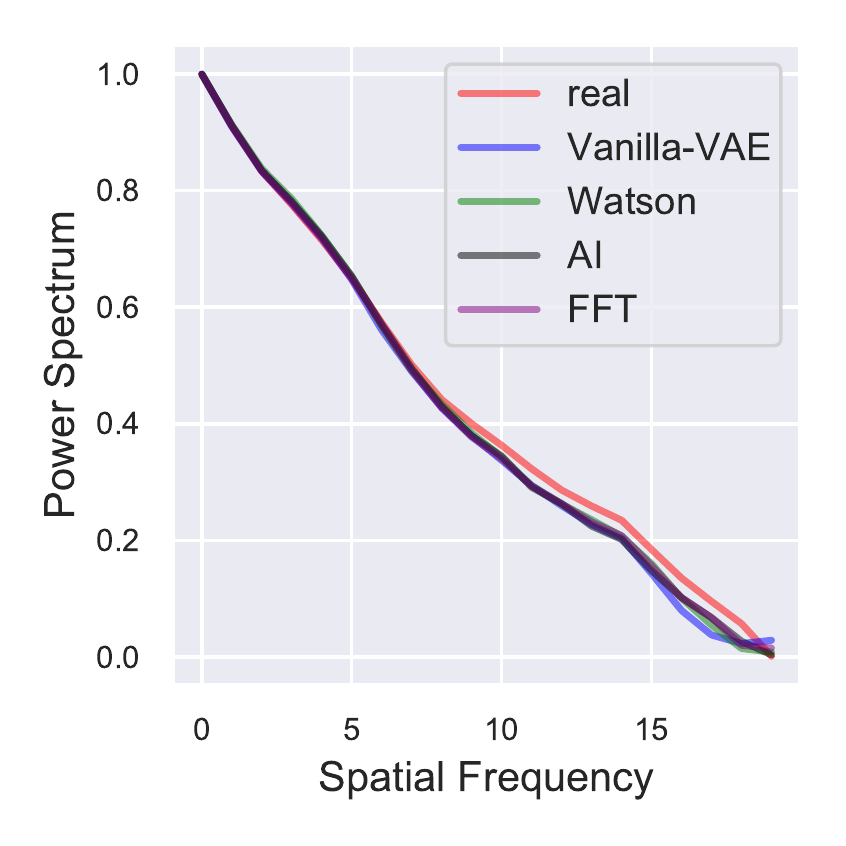}}
\hspace*{\fill}
  \subfloat[CelebA\label{fig:celeba}]{%
        \includegraphics[width=0.5\columnwidth]{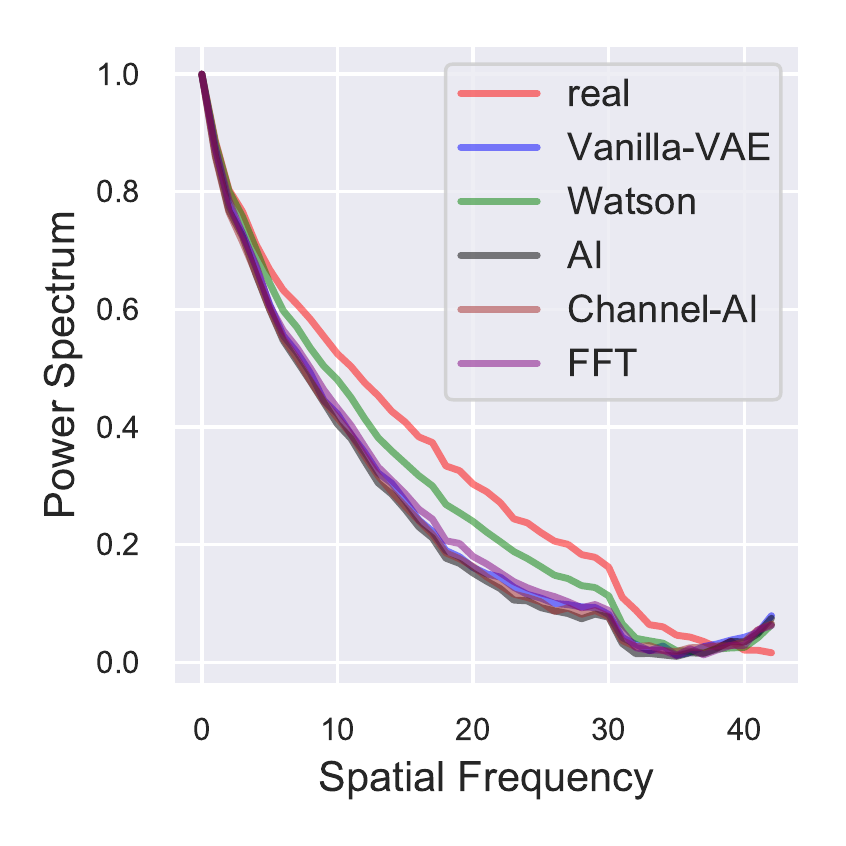}}
\\
\vspace*{-0.12cm}
%%%%%%%%%%%%%%%%%%%%%%%%%%%%%%%%%%%%%%%%%%%%%%%%%%%%%%%%%%%%%%%%%%%%%%%%%%%%%
\subfloat[MNIST+up \label{fig:mnistup}]{%
       \includegraphics[width=0.5\columnwidth]{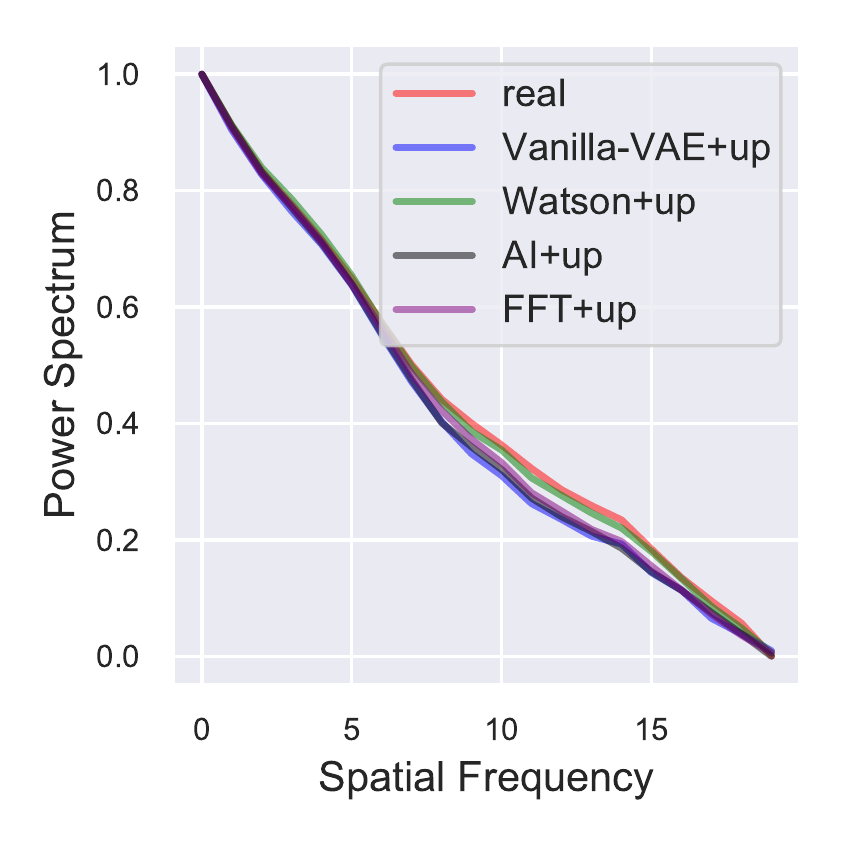}}
\hspace*{\fill}
  \subfloat[CelebA+up\label{fig:celebaup}]{%
        \includegraphics[width=0.5\columnwidth]{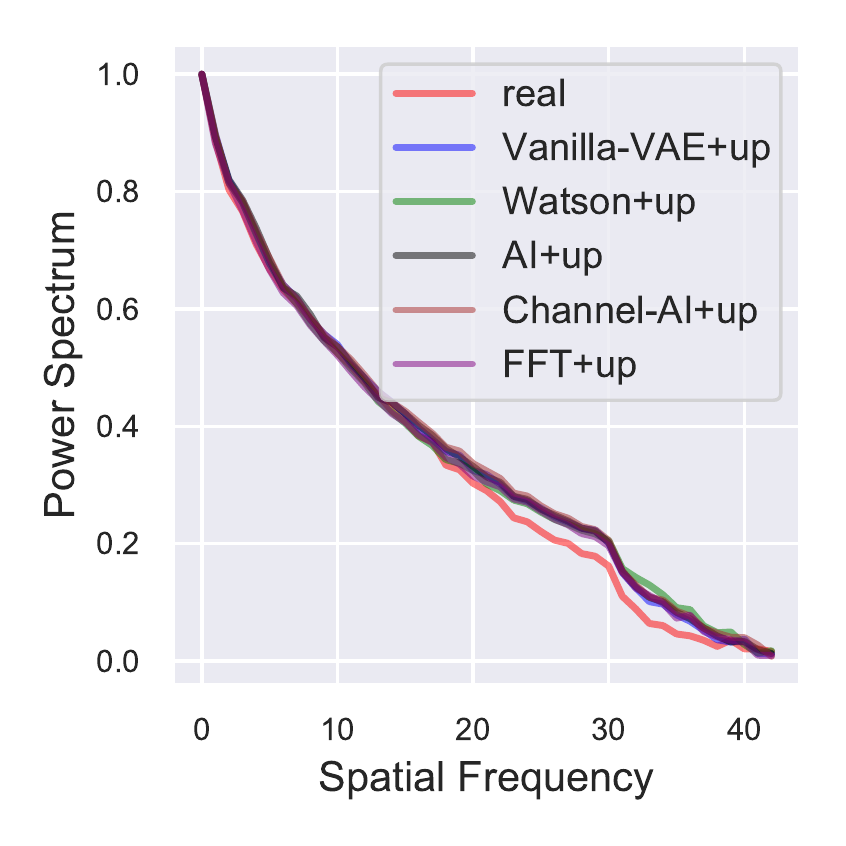}}
\vspace*{-0.15cm}
\caption{Average azimuthal integration power spectrum computed for images in a test batch of either the MNIST \cite{LeCun1998mnist} (first column) or the CelebA dataset \cite{Liu2015celeba} (second column) by applying either the Vanilla-VAE, Watson-DFT, AI or the FFT loss. Results for models trained with the traditional transposed convolution up-sampling operation are shown in \textbf{(a)} and \textbf{(b)}. Corresponding results with the 'N.1.5' up-sampling  \cite{Chandrasegaran2021} are shown in \textbf{(c)}  and \textbf{(d)}.
}
  \label{fig:minmax}
\end{figure} 

\subsection{Spectral regularization with transposed convolution}\label{sec:exp1}
Firstly, we trained VAE models for each of the three different datasets with the baseline Vanilla-VAE, and then with the added Watson-DFT, AI or the FFT loss by employing traditional transposed convolution up-sampling.
Our purpose was to compare the three different ways to achieve SR against each other and the baseline.
Based on this, we evaluated if our contribution, the FFT loss, can compete against the Watson-DFT and the AI loss.
\tabref{tab:exp1} summarizes the quantitative results from the empirical investigation, while \figr{fig:mnist} and \figr{fig:celeba} show the average azimuthal integration of the power spectrum for models trained with the different objective functions.
The quantitative RMSE metrics in \tabref{tab:exp1} show that our proposed FFT loss performs well in generating images that resemble the true images.
The only exception is when the RMSE is computed in the AI-domain, where the Watson-DFT loss has a smaller RMSE than our FFT loss for the CelebA dataset.
This is also shown in \figr{fig:celeba}.

\begin{table}[t]
    \caption{\small Mean $\pm$ std RMSE (lower is better, marked in \textbf{bold}) computed in AI-domain, 2D Fourier Transform (FT) domain and spatial domain, for experiments in \Sec{sec:exp1} for Vanilla-VAE, Watson-DFT loss, AI loss, and the FFT loss with transposed convolution up-sampling.}
    \label{tab:minmax}
    \centering
     \resizebox{1.05\linewidth}{!}{%
  \begin{tabular}{l|l|c|c|c}
 Objective function & Dataset & AI & 2D FT & Spatial domain\\
     \hline
     \hline
    Vanilla-VAE: &  Shape & 1.2226 $\pm$ 0.8317 & 1.1791 $\pm$ 0.3184 & 0.0004 $\pm$ 0.0088\\

   AI: & Shape & 0.8892 $\pm$ 0.4943 & 1.0842 $\pm$ 0.3549 & 0.0003 $\pm$ 0.0055\\ 
   FFT: & Shape & \textbf{0.4447 $\pm$ 0.1671} &\textbf{1.0543 $\pm$ 0.3289} &\textbf{ 0.0001 $\pm$ 0.0021} \\ 
    \hline
           \hline
    Vanilla-VAE: &  MNIST & 3.5091 $\pm$ 0.8296 & 1.7728 $\pm$ 0.1371  & 0.0081 $\pm$ 0.1331\\

    Watson-DFT: & MNIST & 3.9728 $\pm$ 0.9697& 1.8497 $\pm$ 0.0962 & 0.0094 $\pm$ 0.0282\\

   AI: & MNIST & 3.5084 $\pm$ 0.9108 & 1.7274 $\pm$ 0.1194 &0.0079 $\pm$ 0.0228\\ 

   FFT: & MNIST & \textbf{2.8761 $\pm$ 0.5774 }& \textbf{1.5710 $\pm$ 0.1249} & \textbf{0.0062 $\pm$ 0.0174} \\ 
      \hline
              \hline
    Vanilla-VAE: &  CelebA &  9.2300$\pm$ 1.8732 & 4.3506 $\pm$ 0.7480 & 0.0323$\pm$ 0.0376\\

    Watson-DFT: & CelebA & \textbf{6.7433  $\pm$ 2.7038} & 4.0370 $\pm$ 0.7803 & 0.0284 $\pm$ 0.0356 \\

   AI: & CelebA & 9.5100 $\pm$ 2.0860 & 4.3345 $\pm$ 0.6696 & 0.0316 $\pm$ 0.0372\\ 

   FFT: & CelebA & 8.3299 $\pm$ 1.5421 & \textbf{ 3.5406 $\pm$ 0.6743} & \textbf{0.0237 $\pm$ 0.0303}\\ 

\end{tabular}
}
\label{tab:exp1}
\end{table} 

\subsection{Different last-layer up-sampling procedures}\label{sec:exp2}
Secondly, we changed the up-sampling of the last layer of the VAEs from the traditional transposed convolution to 'N.1.5', as introduced in \Sec{sec:up}, and repeated the experiments from \Sec{sec:exp1}.
\tabref{tab:exp2} summarizes the quantitative results from the empirical investigation, while \figr{fig:mnistup} and \figr{fig:celebaup} show average azimuthal integration of the power spectrum for models trained with the Vanilla-VAE, Watson-DFT, AI and FFT loss in combination with 'N.1.5' up-sampling in the last layer. 
In all cases, results in \tabref{tab:exp2} show that models trained with our proposed FFT loss resemble the true data distribution better than any of the other evaluated objective functions.
Comparing the rightmost parts of \figr{fig:mnist} and \figr{fig:celeba} to \figr{fig:mnistup} and \figr{fig:celebaup}, it can be noted that the change in up-sampling to 'N.1.5' improves the alignment of the high frequencies for all generative models.
However, it should be noted that the change in up-sampling does not always imply lower RMSE, compare e.g. Watson-DFT for CelebA in the AI-domain in \tabref{tab:exp1} with the corresponding value in \tabref{tab:exp2}.
For the AI loss and CelebA we can verify the results from \cite{Chandrasegaran2021}; changing the up-sampling to 'N.1.5' reduces the RMSE in both the AI-domain and the 2D Fourier transform-domain.
However, this result is not consistent for all datasets, over all tested SR losses, nor for the baseline Vanilla-VAE.
This indicates that a change in the up-sampling procedure is one possible way to improve the performance of generative models, but the effect is not consistent, and we urge more research on this topic. 
%%%%%%%%%%%%%%%%%%%%%%%%%
\begin{table}[t]
    \caption{\small Mean $\pm$ std RMSE for experiments in \Sec{sec:exp2} for Vanilla-VAE, Watson-DFT loss, AI loss, and the FFT loss, with 'N.1.5' up-sampling in the last layer, following ~\cite{Chandrasegaran2021}.}
    \label{tab:minmax}
    \centering
    \resizebox{1.05\linewidth}{!}{%
  \begin{tabular}{l|l|c|c|c}

 Objective function & Dataset & AI & 2D FT & Spatial domain\\
     \hline
     \hline
    Vanilla-VAE: &  Shape & 1.3893 $\pm$ 0.7328 & 1.1983 $\pm$ 0.3240 & 0.0004 $\pm$ 0.0083\\

   AI: & Shape & 0.8041 $\pm$ 0.4916 & 1.0940 $\pm$ 0.3186 & 0.0002 $\pm$0.0046 \\

   FFT: & Shape &\textbf{ 0.1394 $\pm$ 0.1103} &\textbf{ 1.0325 $\pm$ 0.3248} & \textbf{2.6892e-05 $\pm$ 0.0005}\\
    \hline
    \hline
    Vanilla-VAE: &  MNIST & 3.4711 $\pm$ 1.0887 & 1.8513 $\pm$ 0.1008 & 0.0084 $\pm$0.0234\\

    Watson-DFT: & MNIST & 3.6987 $\pm$ 1.2195& 1.9171 $\pm$0.1164 & 0.0100 $\pm$ 0.0278\\

   AI: & MNIST & 3.1265 $\pm$0.9568 & 1.8270 $\pm$ 0.0848 & 0.0085 $\pm$ 0.0236 \\

   FFT: & MNIST & \textbf{2.9276 $\pm$ 0.9551} & \textbf{1.7041 $\pm$ 0.0958} & \textbf{0.0071 $\pm$ 0.0189} \\
       \hline
       \hline
    Vanilla-VAE: &  CelebA & 9.4452 $\pm$ 3.7583 & 4.3022 $\pm$ 0.7461 & 0.0315 $\pm$ 0.0370\\

    Watson-DFT: & CelebA & 6.8082 $\pm$2.5990 & 4.0807 $\pm$ 0.8466& 0.0290$\pm$ 0.0359\\

   AI: & CelebA & 8.5200 $\pm$4.0584 & 3.6181 $\pm$ 4.3116 & 0.7356 $\pm$ 0.0367 \\

   FFT: & CelebA &\textbf{ 5.8252 $\pm$ 2.5366} & \textbf{3.5917 $\pm$ 0.6925} & \textbf{0.0245 $\pm$ 0.0315} \\
    \hline
\end{tabular}
}
\label{tab:exp2}
\end{table}

%% file: body/conclusion.tex
In this paper, we have shown that a simple spectral regularization term based on the 2D Fourier transform performs better than more complex regularization methods for improving the image quality of the VAE generative model.
Moreover, our results show that changing the up-sampling procedure in the last layer from transposed convolution to nearest-neighbor interpolation followed by standard convolution gives more ambiguous results than indicated by previous research.   
Clearly, more research is needed to untangle the true spectral properties of neural generative models. 